\documentclass[10pt,twocolumn,letterpaper]{article}

\usepackage[pagenumbers]{cvpr} 

\usepackage{times}
\usepackage{epsfig}
\usepackage{graphicx}
\usepackage{amsmath}
\usepackage{amssymb}
\usepackage{epsfig}
\usepackage{graphicx}
\usepackage{amsmath}
\usepackage{amssymb}
\usepackage{todonotes}
\usepackage{lineno}
\usepackage{colortbl}
\usepackage{caption}
\usepackage{subcaption}
\usepackage{booktabs}
\usepackage{xcolor}
\usepackage{soul}
\usepackage{xr}
\usepackage{multirow}
\usepackage{algorithm}
\usepackage{algpseudocode}
\usepackage[accsupp]{axessibility}

\definecolor{darkgreen}{rgb}{0.0, 0.4, 0.0}
\definecolor{orange}{rgb}{1.0, 0.49, 0.0}
\definecolor{purple}{rgb}{0.54, 0.17, 0.89}
\definecolor{ForestGreen}{RGB}{34,139,34}
\definecolor{ModernBlue}{RGB}{0,153,153}
\definecolor{LightModernBlue}{RGB}{0,204,204}
\definecolor{DarkPink}{RGB}{204,0,102}

\newcommand{\datagenerator}{\texttt{M$^{3}$Act}}

\definecolor{cvprblue}{rgb}{0.21,0.49,0.74}
\usepackage[pagebackref,breaklinks,colorlinks,citecolor=cvprblue]{hyperref}

\def\paperID{****} 
\def\confName{CVPR}
\def\confYear{2024}

\title{On the Equivalency, Substitutability, and Flexibility of Synthetic Data}

\author{Che-Jui Chang\\
Rutgers University\\
{\tt\small chejui.chang@rutgers.edu}
\and
Danrui Li\\ 
Rutgers University\\
{\tt\small danrui.li@rutgers.edu}
\and
Seonghyeon Moon\\ 
Rutgers University\\
{\tt\small sm206@cs.rutgers.edu}
\and
Mubbasir Kapadia\\
Roblox\\
{\tt\small mkapadia@roblox.com}
}

\begin{document}

\twocolumn[{%
\renewcommand\twocolumn[1][]{#1}%
\maketitle


}]



\begin{abstract}
%
We study, from an empirical standpoint, the efficacy of synthetic data in real-world scenarios.
Leveraging synthetic data for training perception models has become a key strategy embraced by the community due to its efficiency, scalability, perfect annotations, and low costs.
Despite proven advantages, few studies put their stress on how to efficiently generate synthetic datasets to solve real-world problems and to what extent synthetic data can reduce the effort for real-world data collection.
To answer the questions, we systematically investigate several interesting properties of synthetic data -- the equivalency of synthetic data to real-world data, the substitutability of synthetic data for real data, and the flexibility of synthetic data generators to close up domain gaps.
Leveraging the {\datagenerator} synthetic data generator, we conduct experiments on DanceTrack and MOT17. 
Our results suggest that synthetic data not only enhances model performance but also demonstrates substitutability for real data, with 60\% to 80\% replacement without performance loss. 
In addition, our study of the impact of synthetic data distributions on downstream performance reveals the importance of flexible data generators in narrowing domain gaps for improved model adaptability.

\end{abstract}

\begin{figure*}[t]
\centering
   \includegraphics[width=\linewidth]{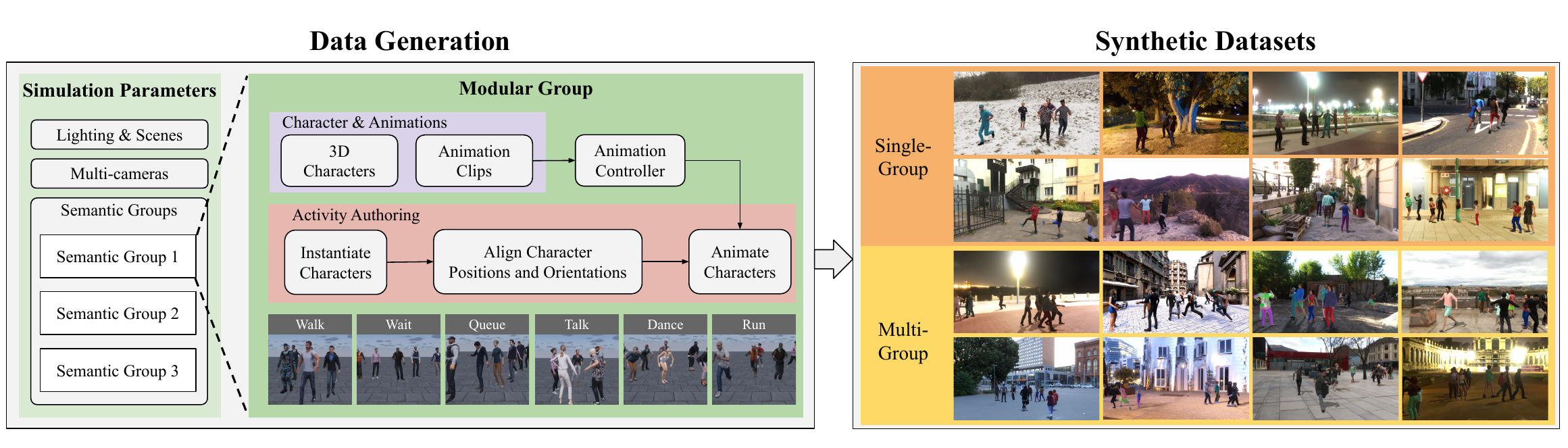}
   \vspace{-15pt}
   \caption{The data generation process of {\datagenerator}. The data generation process is highly parameterized, enabling the adjustment of resulting synthetic data distributions, as illustrated at the right.}
\label{fig:generation_pipeline}
\end{figure*}

\section{Introduction}
For the past decade, collecting real-world data with human annotations has been driving the momentum of research advancements in the field of computer vision and machine perception.
Despite the tremendous breakthrough in many tasks, obtaining these datasets with high-quality human annotations remains costly and labor-intensive, thus hindering the development for certain research tasks that require fine-grained annotations, including human pose and shape estimation\cite{black2023bedlam, ebadi2021peoplesanspeople}, multi-object tracking~\cite{sun2022dancetrack, zhang2023motrv2, zeng2022motr, MeMOTR}, and collective activity understanding~\cite{rahman2022pacmo, cl2021acoustic, chang2022ivi, chang2020transfer, zhou2022composer, chang2023importance, chang2022disentangling, actortransformer}.

The adoption of synthetic data from game engines has become an emerging alternative to real-world data, due to its efficiency, flexibility, scalability, and perfect annotations.
Previous studies \cite{black2023bedlam, yang2023synbody, chang2023learning, varol2017learning, ebadi2021peoplesanspeople} also show the capability of synthetic data in providing diverse and photorealistic images, generating customized datasets for edge cases, improving benchmark performances on downstream datasets, and mitigating ethical risks, for a wide range of vision and perception tasks.
Nonetheless, related studies are rarely focused on the similarity between data distributions, the equivalency of synthetic data to real-world data, the substitutability of synthetic data for real data, and the flexibility of synthetic data generators to close up the domain gap. 
Therefore, it remains unclear to the community how to efficiently generate synthetic datasets to solve real-world problems and to what extent synthetic data can benefit real-world tasks.

In this work, we aim to study the aforementioned properties of synthetic data, by systematically experimenting with different data sources for training and comparing the performance of models trained on synthetic data to those trained on real-world data.
We adopt the multi-person tracking (MPT) task as the focus of our study and leverage {\datagenerator} \cite{chang2023learning} as the synthetic data generator.
{\datagenerator} features multiple semantic groups and produces highly diverse and photorealistic videos with a rich set of annotations suitable for human-centered tasks, including multi-person tracking, group activity recognition, and controllable human group activity generation.
Notably, it offers a flexible data generation pipeline to modify the data distribution by editing the highly parameterized and modularized human groups, enabling us to examine the influence of synthetic data distributions on downstream task performances.

The key contribution of this work lies in the proven efficacies and insights gained from our investigations of the equivalency of synthetic data to real-world data and the substitutability of synthetic data for real data. 
We show that synthetic data can not only improve model performance on downstream datasets but also effectively replace up to 80\% of MOT17 data \cite{milan2016mot16} without compromising performances.
Moreover, our investigation of the impact of synthetic data distributions on downstream model performances highlights the importance of a flexible synthetic data generator to shorten domain gaps, thereby enhancing the adaptability of models trained on synthetic data to real-world scenarios

\section{Background}


Synthetic data plays a vital role in contemporary computer vision and machine learning research, particularly because of the increasing demand driven by large data-hungry models \cite{videoworldsimulators2024, rombach2022high}. 
The community widely acknowledges the benefits of employing synthetic data for training machine learning models, due to its efficiency, enhanced performance, customizability, and cost reduction.
A key characteristic of synthetic data is its task-specific nature, where datasets are tailored to specific research objectives or scenarios. 
This customization allows researchers to design data generators to produce synthetic data that closely aligns with the requirements of the downstream experiments, enhancing the performance of machine learning models trained on synthetic data. 
For example, SURREAL~\cite{varol2017learning} was designed for human pose estimation to improve the background and data diversity.
BEDLAM~\cite{black2023bedlam} integrates physically animated clothes into its data generation pipeline for improved performance on human pose and shape estimation.
{\datagenerator}~\cite{chang2023learning} leverages the rule-based authoring of human group activities for multi-person and multi-group research. 
The same study also shows that datasets such as BEDLAM~\cite{black2023bedlam} and GTA-Humans~\cite{cai2021playing} are inadequate for multi-person tracking. 

Despite these advantages, prior studies have primarily focused on the enhanced performances achieved with synthetic data. 
While performance enhancement is crucial, seeking optimality may not always be practical in real-world scenarios. 
Understanding the tradeoffs involved in collecting more data and the extent to which it leads to performance improvements is equally important.
Previous studies \cite{chang2023learning} have hinted at the potential for synthetic data to replace certain amounts of real data, yet these properties of synthetic data have not been thoroughly explored. 
In this study, we delve into several interesting properties -- the equivalency, substitutability, and flexibility of synthetic data. 
We demonstrate through experiments how these properties can provide valuable insights for using synthetic data in real-world scenarios.

\section{Overview of {\datagenerator}}
We provide an overview of the data generation process, the flexible parameterization, and the properties of {\datagenerator}.

\subsection{Synthetic Data Generation}
{\datagenerator} is a synthetic data generator built with
Unity Engine~\cite{borkman2021unity}. 
It features multiple semantic groups and produces highly diverse and photorealistic videos with a rich set of annotations suitable for human-centered tasks.
The data generator contains 25 scenes, 104 HDRIs,
5 lighting volumes, 2200 human models, 384 animations, and 6 group activities.
{\datagenerator} is tailored to support multi-person and multi-group research, which effectively enhances the model performances on multi-person tracking and group activity recognition, and enables novel research, controllable group activity generation.
Its data generation process, illustrated in Fig.~\ref{fig:generation_pipeline}, is procedurally driven and contains a high degree of simulation parameters, including scene parameters like lighting, scenes, and cameras, as well as group parameters such as animations, characters, alignments, and group types. 
This level of flexibility and customization is rarely seen in previous synthetic data works \cite{black2023bedlam, yang2023synbody, varol2017learning, bazavan2021hspace}. 
It allows for the adjustment of the simulation parameters within the generator, enabling the customization of synthetic data distributions to match specific research requirements.

\subsection{Implications}
Despite the promising contributions of {\datagenerator} to performance enhancement, their study has revealed initial findings on several interesting properties of synthetic data.
First, the variability in synthetic data distributions resulting from the flexible parameterization of {\datagenerator} presents a unique challenge in ensuring the suitability of synthetic data for downstream tasks.
This challenge is particularly pronounced in multi-person scenarios, which are significantly more complex compared to tasks addressed by previous synthetic datasets \cite{black2023bedlam, yang2023synbody, cai2021playing}.
Second, synthetic data generated by {\datagenerator} exhibits properties of equivalency and substitutability, with results indicating that synthetic data can effectively replace a substantial portion of real data for multi-person tracking. 
Our deep investigation of these properties extends beyond the initial findings and offers valuable insights into the understanding of these synthetic data properties in real-world scenarios.

\section{Experiments}
We evaluate the effectiveness of synthetic data by conducting experiments on multi-person tracking (MPT),
the objective of which is to predict the tracklets of all individual persons given an input video stream.
The tracking performances are measured on two distinct real-world datasets: DanceTrack~\cite{sun2022dancetrack} (DT) and MOT17~\cite{milan2016mot16}.
The former presents a particularly challenging multi-person tracking scenario characterized by dynamic dance movements with individuals of uniform outfits.
It has a total of 100 videos with over 105K frames.
The latter is a widely used dataset for multi-object tracking, with objects including pedestrians, bicycles, and cars, captured in outdoor scenes. 
The main challenges involve crowded scenarios and interruptions in tracking due to dynamic camera movements. 
The dataset contains 7 long videos with a total of 11K frames.

\subsection{Preliminaries}
Our experiments cover the following properties of synthetic data and research questions:
\begin{itemize}
    \item \textit{Equivalency} and \textit{Substitutability}. How much synthetic data is equivalent to real data? In other words, how much real data can be substituted by synthetic data, without sacrificing performance? Answers to the question would help the community understand the practical merits of using synthetic data in reducing the cost of data collection and annotation in real-world scenarios.
    \item \textit{Distribution} and \textit{Flexibility}. How does the distribution of the synthetic data affect the performance of target datasets and can we narrow the domain gaps by adjusting the distribution of synthetic data?
    This helps us understand the protocol of designing and selecting useful synthetic data for enhancing downstream task performance in the target domain.
\end{itemize}

\begin{table}[t]
\setlength{\tabcolsep}{2.4pt}
\small
  \aboverulesep=0ex
  \belowrulesep=0ex 
    \begin{center}
    \begin{tabular}{l|c|c|c|c|c}
        \toprule
        Pretrain $\rightarrow$ Finetune & HOTA$\uparrow$ & DetA$\uparrow$ & AssA$\uparrow$ & IDF1$\uparrow$ & MOTA$\uparrow$ \\
        \midrule
        N/A $\rightarrow$ 100\% MOT17 & 56.3 & 51.9 & 62.1 & 66.6 & 56.1 \\
        \midrule
        Syn $\rightarrow$ 20\%  MOT17 & 51.7 & 48.0 & 56.3 & 62.1 & 54.4 \\
        Syn $\rightarrow$ 40\%  MOT17 & 57.4 & 54.3 & 61.5 & 69.3 & 61.3 \\
        Syn $\rightarrow$ 60\%  MOT17 & 59.7 & 57.1	& 63.0 & 70.0 &	62.9 \\
        Syn $\rightarrow$ 80\%  MOT17 & 60.6 & 58.1 & 63.8 & 73.0 & 67.1\\
        Syn $\rightarrow$ 100\% MOT17 & \textbf{63.7} & \textbf{61.6} & \textbf{66.9} & \textbf{74.7} & \textbf{68.7}\\
        \bottomrule
    \end{tabular}
    \end{center}
    \vspace{-10pt}
    \caption{Tracking results on MOT17 with different amounts of real data.}
    \label{tab:mot_result}   
\end{table} 

\begin{figure}[t]
\centering
   \includegraphics[width=0.75\linewidth]{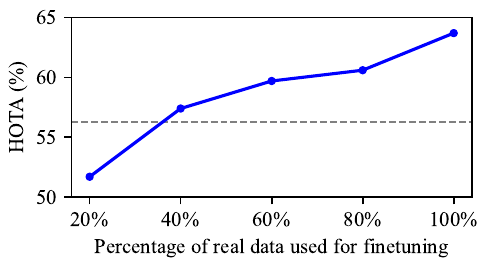}
   \vspace{-10pt}
   \caption{Plot showing tracking performances with different percentages of MOT17 data for fine-tuning. The dotted line represents the performance when the model is trained solely on 100\% real data. }
\label{fig:mot_result}
\end{figure}

To answer the first question, we conduct the experiment on MOT17~\cite{milan2016mot16} dataset.
We first train the model with 100\% real data as a comparison baseline.
Then we use a constant amount of synthetic data for pretraining and then gradually reduce the amount of real data used for finetuning. 
This enables us to not only observe the changes in tracking performances but also identify how much real data can be replaced by synthetic data, by comparing the performances with the baseline.
We follow the protocol of {\datagenerator}~\cite{chang2023learning} and use the same portion of synthetic data in this experiment. 
For fair comparisons, we pre-trained the model on synthetic data for 5 epochs and then fine-tuned it for 20 epochs for all conditions.

For the second research question, we extend the original MPT experiments on DanceTrack~\cite{sun2022dancetrack} in {\datagenerator} by adjusting the synthetic data distributions. 
Specifically, we use the {\datagenerator} data generator to prepare various datasets with different combinations of human groups.
A total of 5 synthetic datasets are constructed, including 1K video clips of a single ``Dance'' group (``D''), 1.5K videos with a “walk” group and a “run” group simulated at the same time (denoted as  ``WalkRun'' or ``WR''),  2.5K videos of ``WR''+``D'',  6K videos of all ``Single-group'' activities, and 9K videos of all ``Multi-group'' activities (simulated simultaneously).
We then investigate the influence of these synthetic data distributions on model performance in the downstream target dataset. 



\subsection{Benchmark Method}
Recent studies \cite{zeng2022motr, zhang2023motrv2, MeMOTR, yan2023bridging} have demonstrated the superior effectiveness of end-to-end methods over traditional tracking-by-detection approaches \cite{bewley2016simple, cao2023observation, wojke2017simple}, particularly evident in challenging datasets like DanceTrack \cite{sun2022dancetrack} and MOT \cite{milan2016mot16, dendorfer2020mot20}. 
Therefore, our study aims to showcase the effectiveness of synthetic data in improving downstream task performance using end-to-end tracking models, which offer a better assessment of impact of synthetic data on overall tracking performance. 
We primarily evaluate the performance using MOTRv2 \cite{zhang2023motrv2}, the state-of-the-art method across various benchmark datasets. 
MOTRv2 is an extension of MOTR \cite{zeng2022motr} by integrating YOLO-X \cite{ge2021yolox} for enhanced detection bootstrapping.


\subsection{Results}

Tab.~\ref{tab:mot_result} presents the tracking results on MOT17~\cite{milan2016mot16}. 
First, pretraining with our synthetic data leads to significant performance gain on all five metrics, compared to the model without pretraining. 
This observation aligns with several previous studies \cite{black2023bedlam, chang2023learning, yang2023synbody, varol2017learning} that adding synthetic data improves model performances.
Second, the performance improves with the increased amount of real data, as illustrated in Fig.~\ref{fig:mot_result}. 
The model finetuned on only 20\% (or less) real data is inferior to the same model trained on 100\% real data, which suggests a domain gap between the synthetic and real data.
Last, the model trained solely on 100\% real data achieves comparable performance to the model finetuned on 20\% to 40\% of real data.
In other words, \textit{our synthetic data is equivalent to 60\% to 80\% of MOT17 training data.}

\begin{table}[t]
\setlength{\tabcolsep}{2.1pt}
\renewcommand{\arraystretch}{0.95}
\small
  \aboverulesep=0ex
  \belowrulesep=0ex 
    \begin{center}
    \begin{tabular}{l|c|c|c|c|c}
        \toprule
        Training Data & HOTA$\uparrow$ & DetA$\uparrow$ & AssA$\uparrow$ & IDF1$\uparrow$ & MOTA$\uparrow$ \\
        \midrule
        \footnotesize{DT} & 68.8 & 82.5 & 57.4 & 70.3 & 90.8  \\
        \footnotesize{DT + Syn (D)} & 59.0 & 75.5 & 46.1 & 59.0 & 82.6 \\
        \footnotesize{DT + Syn (WR)} & 70.1 & 83.1 & 59.4 & 72.5 & 92.0 \\
        \footnotesize{DT + Syn (WR+D)} & \textbf{71.9} & \textbf{83.6} & \textbf{62.0} & \textbf{74.7} & \textbf{92.6} \\
        \midrule
        \footnotesize{DT + Syn (Single-group)}  & 65.1 & 80.2 & 55.8 & 66.7 & 89.1 \\
        \footnotesize{DT + Syn (Multi-group)}  & \textbf{73.5} & \textbf{83.9} & \textbf{64.7} & \textbf{75.8} & \textbf{93.0} \\
        \bottomrule
    \end{tabular}
    \end{center}
    \vspace{-10pt}
    \caption{MPT results on DanceTrack under different synthetic data distributions (eg. group types). We use all single-group and multi-group data for the last two rows. 
    }
    \label{tab:synthetic_data_type}   
\end{table}

We present in Tab.~\ref{tab:synthetic_data_type} the results on DanceTrack using various group types of synthetic data for training. 
These results reveal that training with multi-group synthetic data, such as "WR" and "Multi-group," leads to superior performance. 
The design of multi-group synthetic data, with multiple human activities animated in the same scene, provides frequent identity switches and produces high-complexity datasets that are suitable for challenging target datasets with dynamic movements such as DanceTrack.
On the contrary, using synthetic data with each subject animated procedually and independently, such as "D" and "Single-group", yields inferior downstream performance to the baseline that is trained without any synthetic data. 
Our findings suggest that \textit{synthetic datasets with apparent similarities in data complexity and distribution to target datasets tend to enhance model performance.} 
It also underscores the importance of having a flexible data generator, such as the modular and highly parameterized groups in {\datagenerator}, which allows for the customization of synthetic data distributions to closely approximate those of target real-world datasets.

\subsection{Discussions}

We discuss the following properties of synthetic data in our study and how the experimental results help to answer the research questions and provide insights for the potential implications of synthetic data.

\subsubsection{Equivalency and Substitutability}

Synthetic data has shown the potential to replace a substantial portion of real-world data, as evidenced by previous studies \cite{chang2023learning, black2023bedlam}. For instance, results from \cite{chang2023learning} indicate that synthetic data generated by {\datagenerator} can substitute for 62.5\% more real data from DanceTrack. 
In our study, we found that synthetic data can replace 60\% to 80\% of real-world data without sacrificing performance on MOT17. 
When considering the total number of image frames in both datasets, the equivalency ratio of synthetic to real data is approximately 30 times. 
Similarly, when considering the total number of track frames (or annotated bounding boxes), the ratio is roughly 5.6 times. 
These equivalency measurements provide valuable insights into the suitability and efficiency of different synthetic datasets for target downstream tasks. 
Furthermore, our results highlight that, despite existing domain gaps between synthetic and real datasets, the substitutability of synthetic data for target real-world data can significantly reduce data collection and annotation costs.

\subsubsection{Impact of Synthetic Data Distributions}
Although synthetic data offers practical benefits such as enhanced performance, substitutability, and cost reduction, its effectiveness relies on how closely its distribution matches that of the target real-world data.
Our experiments suggest that not all synthetic datasets contribute equally to enhancing downstream task performances, even when designed for the same task. 
In fact, adding some synthetic data to the training dataset could even worsen performance. 
While most synthetic data generators \cite{varol2017learning, yang2023synbody} offer limited parameterization options
during the generation process, more complex tasks like multi-person tracking or scenarios involving multiple groups necessitate finer adjustments to the configuration parameters such as group types, person alignments, and group size within the data generator.
Therefore, providing flexibility in synthetic data generators and adjusting parameters to generate synthetic data are crucial for narrowing domain gaps and ensuring the efficacy of using synthetic data in real-world applications.



\section{Conclusion}

In this study, we explore the equivalency, substitutability, and flexibility of synthetic data, offering insights into its practical applications in real-world scenarios. 
Despite our efforts and promising results shown in the experiments, concrete evidence of complete substitution of synthetic data for real data in training datasets remains elusive. 
One significant challenge arises from the existence of domain gaps between synthetic and real data, which ultimately limit the adaptability of models trained solely on synthetic data. 
As we move forward, addressing these limitations will be crucial in maximizing the potential of synthetic data and enhancing its utility in various machine learning tasks.

{
    \small
    \bibliographystyle{ieeenat_fullname}
    \bibliography{main}
}


\end{document}